\documentclass[12pt]{article}

\usepackage{hyperref}
\hypersetup{
    colorlinks=true, 
    linkcolor=blue,  
    urlcolor=blue,   
    citecolor=blue   
}

\usepackage{times}
\usepackage{graphicx}
\usepackage{adjustbox}
\usepackage{booktabs}
\usepackage{floatpag}
\usepackage{multirow}
\usepackage{xcolor,colortbl}
\usepackage{xspace}
\usepackage{alias}
\usepackage{xcolor}
\usepackage{amsmath}
\usepackage{float}
\usepackage{amssymb}

\DeclareMathOperator*{\argmin}{arg\,min}

\newenvironment{sciabstract}{%
\begin{quote} \bf}
{\end{quote}}

\title{GOAT: GO to Any Thing} 

\author{
Matthew Chang,$^{\ast1}$ Theophile Gervet,$^{\ast2}$ Mukul Khanna,$^{\ast3}$ Sriram Yenamandra,$^{\ast3}$\\
Dhruv Shah,$^{4}$, So Yeon Min,$^{2}$, Kavit Shah,$^{5}$, Chris Paxton,$^{5}$ Saurabh Gupta,$^{1}$\\
Dhruv Batra,$^{5}$ Roozbeh Mottaghi,$^{5}$ Jitendra Malik,$^{\ast4,5}$ Devendra Singh Chaplot$^{\ast6}$\\
\normalsize{$^\ast$Equal Contribution,}\\
\normalsize{$^{1}$University of Illinois Urbana-Champaign,} \normalsize{$^{2}$Carnegie Mellon University,}\\
\normalsize{$^{3}$Georgia Institute of Technology,} \normalsize{$^{4}$University of California, Berkeley,}\\
\normalsize{$^{5}$Meta AI Research},
\normalsize{$^{6}$Mistral AI}\\
\normalsize{\href{https://theophilegervet.github.io/projects/goat}{Project Website}}
}

\date{}

\begin{document} 
\maketitle 

\begin{sciabstract}
In deployment scenarios such as homes and warehouses, mobile robots are expected to autonomously navigate for extended periods, seamlessly executing tasks articulated in terms that are intuitively understandable by human operators.
We present GO To Any Thing (GOAT), a universal navigation system capable of tackling these requirements with three key features: a) Multimodal: it can tackle goals specified via category labels, target images, and language descriptions, b) Lifelong: it benefits from its past experience in the same environment, and c) Platform Agnostic: it can be quickly deployed on robots with different embodiments. 
GOAT is made possible through a modular system design and a continually augmented instance-aware semantic memory that keeps track of the appearance of objects from different viewpoints in addition to category-level semantics.
This enables GOAT to distinguish between different instances of the same category to enable navigation to targets specified by images and language descriptions. \\
In experimental comparisons spanning over 90 hours in 9 different homes consisting of 675 goals selected across 200+ different object instances, we find GOAT achieves an overall success rate of 83\%, surpassing previous methods and ablations by 32\% (absolute improvement). 
GOAT improves with experience in the environment, from a 60\% success rate at the first goal to a 90\% success after exploration. 
In addition, we demonstrate that GOAT can readily be applied to downstream tasks such as pick and place and social navigation.
\end{sciabstract}

\section{Introduction}


Ever since there were animals that could move, navigation to desired locations – food, mates, nests - has been a fundamental aspect of animal and human behavior. The scientific study of navigation is a very interdisciplinary field to which contributions have been made by, among others, researchers in ethology, zoology, psychology, neuroscience, and robotics. In this paper, we present a mobile robot system that was inspired by some of the most salient findings from animal and human navigation.
\begin{itemize}
\item {\it Cognitive maps.} Many animals maintain internal spatial representations of their environment. There has been a vigorous debate on the nature of this map – is it metric in a Euclidean sense or just topological- and in Nobel prize-winning work, neural correlates of cognitive maps have been found in the hippocampus. This suggests that a purely reactive, memoryless navigation system is inadequate for robotics.~\cite{tolman1948cognitive}
\item {\it How is this internal spatial representation acquired?}  From human studies, it has been argued that these are built up through “route-based” knowledge. In the process of daily or other episodic activities,  we learn the structure of a route – origin, destination, waypoints etc. Over time features from different experiences are integrated together into a single layout representation, the “map”. For mobile robots, this motivates a version of ``lifelong learning" - continual improvement of the internal spatial representation as the mobile robot performs active search and exploration.~\cite{hilton2023route}
\item {\it Is navigation driven exclusively by the geometric configuration of locations?} No, as the visual appearance of landmarks plays a major role in animal and human navigation. This suggests maintaining rich multimodal representations of the spatial environment of the mobile robot.~\cite{collett2002memory}
\end{itemize}

\begin{figure}[t!]
    \thisfloatpagestyle{empty}
    \centering
    \includegraphics[width=\textwidth,keepaspectratio]{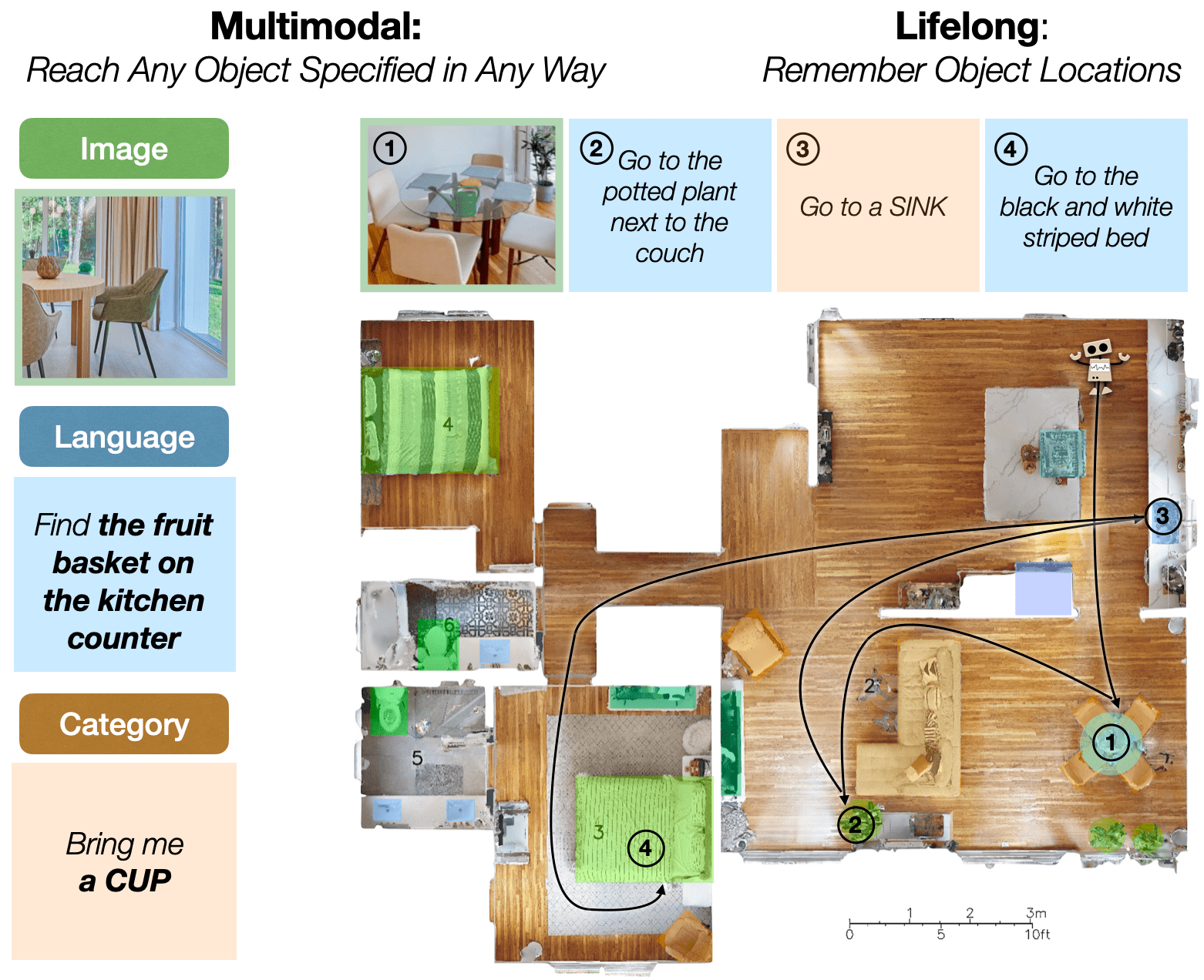}
    \caption{
    \small
    \textbf{GOAT (GO to Any Thing) task.} 
    The GOAT task requires lifelong learning, meaning taking advantage of past experience in the same environment, for multimodal navigation. 
    The robot must be able to reach any object specified in any way and remember object locations to come back to them.
    }
    \label{fig:goat_task}
\end{figure}

Let us get concrete. Consider a robot starting in an unseen environment as shown in Figure~\ref{fig:goat_task}, and suppose it is asked to find a dining table image (goal 1). Navigating to this goal requires recognizing that the picture shows a dining table and having the semantic understanding of indoor spaces to efficiently explore the home (e.g. dining tables are not found in the bathroom).
Suppose the robot is then asked to \textit{Go to the potted plant next to the couch} (goal 2). This requires visual grounding of the text instruction in the physical space. The next instruction is to \textit{Go to a SINK} (goal 3), the capitalization emphasizing that any object of the category SINK is a valid goal. In this example, the robot has already seen a sink in the house during the first task, so it should remember its location and be able to plan a path to reach it efficiently. This requires the robot to build, maintain and update a lifelong memory of the objects in the environment, their visual and linguistic properties and their latest location. Given any new multimodal goal, the robot should also be able to query the memory to determine whether the goal object already exists in the memory or requires further exploration. In addition to these capabilities for multimodal perception, exploration, lifelong memory, and goal localization, the robot also needs effective planning and control to reach the goal while avoiding obstacles.

In this paper, we present \textit{GO to Any Thing} (GOAT), a universal navigation system with three key features: 
a) \textbf{Multimodal:} it can tackle goals specified via category labels, target images, and language descriptions,
b) \textbf{Lifelong:} it benefits from its past experience in the same environment in the form of a map of objects instances (as opposed to stored implicitly within the parameters of a machine learning model) updated over time, and 
c) \textbf{Platform Agnostic:} it can be seamlessly deployed on robots with different embodiments — we deploy GOAT on a quadruped and a wheeled robot.
GOAT is made possible through the design of an {\it instance-aware semantic memory} that keeps track of
the appearance of objects from different viewpoints in addition to category-level semantics. This enables GOAT to distinguish between
different instances of the same category to enable navigation to targets
specified by images and fine-grained language descriptions. This memory is
continually augmented as the agent spends more time in the environment, leading
to improved efficiency in reaching goals over time.  

In experimental comparisons spanning over 90 hours in 9 different homes consisting of 675 goals selected across 200+ different object instances, we find GOAT achieves an overall success rate of 83\%, surpassing previous methods and ablations by 32\% (absolute improvement). 
GOAT performance improves with experience in the environment from a 60\% success rate at the first goal to 90\% success rate once the environment is fully explored.
In addition, we demonstrate that GOAT, as a general navigation primitive, can readily be applied to downstream tasks like pick and place and social navigation.
GOAT’s performance can in part be attributed to the modular nature of the system: it leverages learning in the components in which it is required (\ie object detection, image/language matching) while still leveraging strong classical methods (\ie mapping and planning). Modularity is also responsible for the ease of deployment across different robot embodiments and downstream applications, as individual components can be easily adapted or new components introduced.

While there is a large body of work on navigation~\cite{10.5555/1121596}, most only evaluate in simulation or develop specialized solutions to tackle a subset of these tasks. 
Classical robotics works~\cite{thrun1999minerva} employed geometric reasoning to solve navigation to geometric goals.
With advances in semantic understanding of images,
researchers started using semantic reasoning to improve exploration efficiency in novel 
environments~\cite{chaplot2020learning} and tackling semantic goals specified via 
categories~\cite{ramakrishnan2022poni, gupta2017cognitive, anderson2018vision, chaplot2020object, krantz2020beyond, shridhar2020alfred, chang2020semantic}, 
images~\cite{zhu2017target, chaplot2020neural, hahn2021no, krantz2022instance, krantz2023navigating} and language instructions~\cite{min2021film, song2023llm, gadre2023cows}. 
Most of these methods
are a) specialized to a single task (\ie they are uni-modal), b) only tackle a single goal in each 
episode (\ie are not lifelong), and c) evaluated only in simulation (or rudimentary real-world environments).
GOAT advances upon these works on all three fronts and tackles
multiple goal specifications in a lifelong manner in the real world.
This supersedes past works that only innovate along one axis, \eg 
past works~\cite{wani2020multion, chaplot2020neural} 
tackle a sequence of goal but goals are limited to either be object goals~\cite{wani2020multion} or image goals~\cite{chaplot2020neural} in simulation, 
\cite{al2022zero} tackle flexible goal specifications but only show simulated results for one goal per episode, and 
\cite{gervet2023navigating} show real world results but only for reaching one object goal per episode.
Inspired by animal and human navigation, GOAT maintains a map of the environment as well as visual landmarks - egocentric views of object instances - which are stored in our novel instance-aware object memory.
This memory should be queryable with both images and natural language to satisfy GOAT’s multimodality requirement.
We enable this by storing raw images for visual landmarks, as opposed to features, allowing us to leverage recent advances in image-image matching and image-language matching independently.
We use Contrastive Language-Image Pretraining (CLIP)~\cite{radford2021learning} for image-language matching and SuperGlue~\cite{sarlin2020superglue} for image-image matching.
CLIP follows a long history of associating text with images or regions in images~\cite{hernandez2021natural,farhadi2010every, frome2013devise, fang2015captions, johnson2016densecap, krishna2017visual, plummer2015flickr30k} and has led to the development of language-conditioned open-vocabulary object detectors~\cite{zhou2022detecting,lseg,owlvit}. 
CLIP itself, or object detectors derived from CLIP
have recently been used for robotic tasks, \eg object search~\cite{gadre2023cows}, mobile manipulation~\cite{ovmm}, and table-top manipulation~\cite{shridhar2022cliport}. 
Similarly, SuperGlue follows a long history of geometric image matching~\cite{huttenlocher1990recognizing, lowe2004distinctive} with recent learning-based methods~\cite{sarlin2020superglue} leading to better performance in certain situations. 
Recent work has started evaluating these in embodied settings where a robot must navigate either to an image in the world~\cite{krantz2022instance,chaplot2020neural} or to an image corresponding to a particular object instance~\cite{krantz2023navigating}.

GOAT's memory representation follows a long history of scene representation in robotics over the last 40 years: occupancy maps (with geometry~\cite{elfes1989using}, explicit semantics~\cite{salas2013slam++,chaplot2021seal}, or implicit semantics~\cite{gupta2017cognitive}), topological representations~\cite{chaplot2020neural, choset2001topological, kuipers1991robot, savinov2018semi}, and neural feature fields~\cite{simeonov2022neural,shafiullah2022clip,marza2023autonerf,bolte2023usa}. 
Many of these works have started using pre-trained vision-language features like CLIP~\cite{radford2021learning} and either projecting them into 3D directly~\cite{jatavallabhula2023conceptfusion} or capturing them in an implicit neural field~\cite{shafiullah2022clip,bolte2023usa}.
Parametric representations summarize the environment into low-dimensional abstract features, while non-parametric representations view the collection of images itself as a representation. 
Our work leverages aspects of both. 
We build a semantic map for navigating to objects but also store raw images associated with discovered objects (landmarks).

\section{Results}
\begin{figure}
    \thisfloatpagestyle{empty}
    \centering
    \includegraphics[width=0.95\textwidth,keepaspectratio]{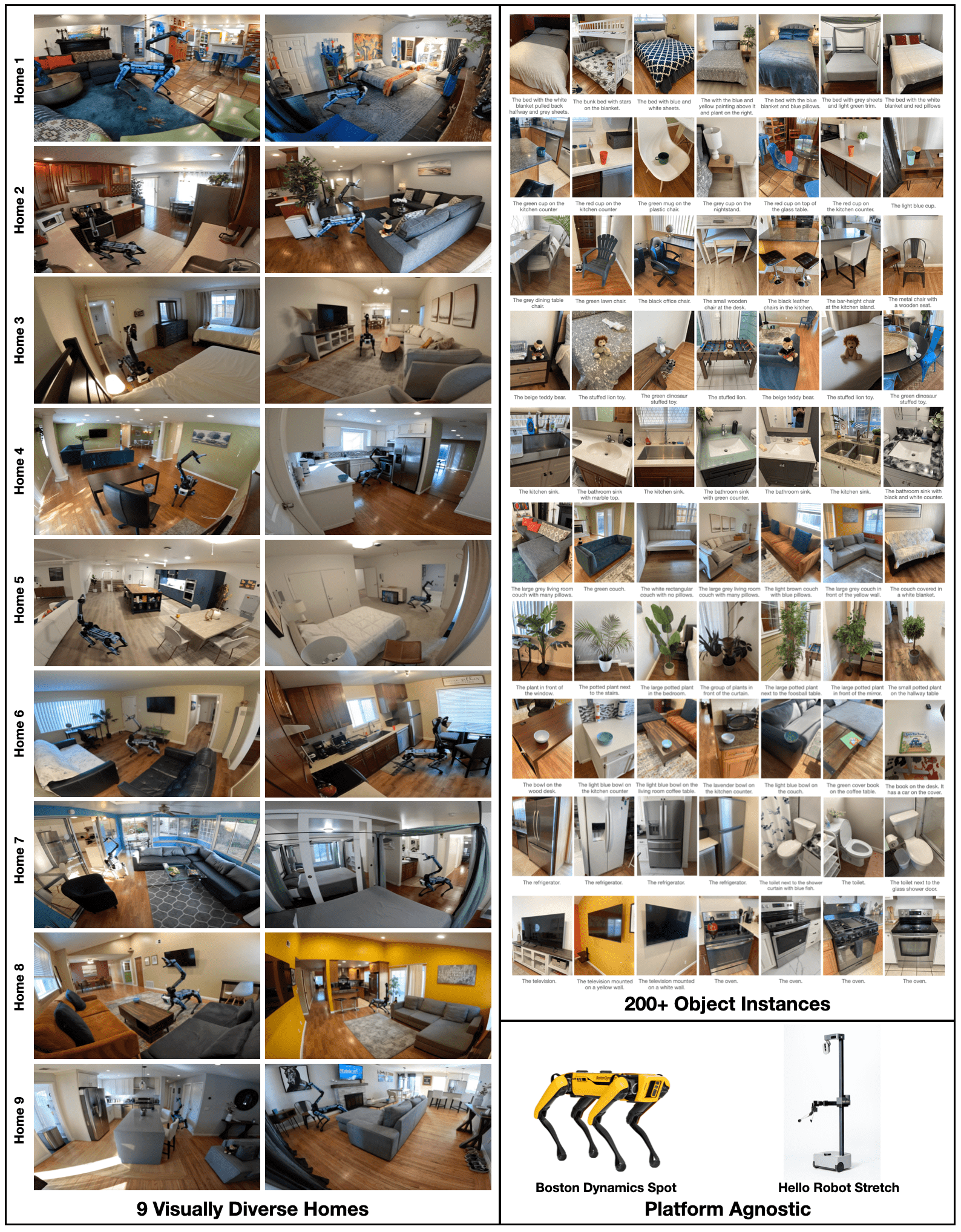}
    \caption{
    \small
    \textbf{``In-the-wild" evaluation.} We deploy the GOAT navigation policy in 9 visually diverse homes and evaluate in on reaching 200+ different object instances as category, image, or language goals. GOAT is platform-agnostic: we deploy it on both Boston Dynamics Spot and Hello Robot Stretch.
    }
    \label{fig:environment_diversity}
\end{figure}

Video 1 summarizes our results. We deployed GOAT on and conducted qualitative experiments with Boston Dynamics Spot and Hello Robot Stretch robots. We conducted large-scale quantitative experiments with GOAT on Spot (due to its higher reliability) against 3 baselines in 9 real-world homes to reach a total of 200+ different object instances (see Figure ~\ref{fig:environment_diversity}).

\subsection{Go To Any Thing: Lifelong Learning for Multimodal Navigation}
\label{sec:online}

We formalize the Go to Any Thing task $T$ as follows. We construct navigation episodes consisting of a sequence of unseen goal objects to be reached in unseen environments. The robot is spawned at a random location. At every timestep $t$, the robot receives observations consisting of an RGB image $I_t$, depth image $D_t$, and pose reading $x_t$ from onboard sensors, as well as the current object goal $g_k$, $k \in \{1,2,..,5-10\}$, which consists in an object category (\emph{SINK}, \emph{CHAIR}), an image or language description (\emph{the potted plant next to the couch}, \emph{the black and white striped bed}) uniquely identifying an object instance in the environment. The robot must reach the goal object $g_k$ as efficiently as possible within a limited time budget. As soon as it reaches the current goal or when the time budget is exhausted, the robot receives the next goal to navigate to, $g_{k+1}$. In searching for this sequence of goals the agent is allowed to maintain a memory computed using incoming observations. In this way, if $g_{k+1}$ has been observed during the process of reaching $g_k$ the agent can often more efficiently navigate to $g_{k+1}$.

\subsection{Navigation Performance in Unseen Natural Home Environments}

In this section, we evaluate the ability of the GOAT agent to tackle the GOAT task, i.e., reach a sequence of unseen multimodal object instances in unseen environments.

\paragraph{GOAT Agent} Figure \ref{fig:goat_architecture1} (A) shows an overview of the GOAT system. As the agent moves through the scene, the perception system processes RGB-D camera inputs to detect object instances and localize them into a top-down semantic map of the scene. In addition to the semantic map, GOAT maintains an Object Instance Memory that localizes individual instances of object categories in the map and stores images in which each instance has been viewed. This Object Instance Memory gives GOAT the ability to perform lifelong learning for multimodal navigation. When a new goal is specified to the agent, a global policy first searches the Object Instance Memory to see if the goal has already been observed. After an instance is selected, its stored location in the map is used as a long-term point navigation goal. If no instance is localized, the global policy outputs an exploration goal. A local policy finally computes actions towards the long-term goal. We will dive into more details in the Materials and Methods section.

\paragraph{Instance Matching Strategy} The matching module of the global policy has to identify the goal object instance among previously seen object instances in the Object Instance Memory. We evaluated different design choices and settled on the following: match language goal descriptions with object views in memory using the cosine similarity score between their CLIP~\cite{radford2021learning} features, match image goals with object views in memory using keypoint-based matching with SuperGLUE~\cite{sarlin2020superglue}, represent object views in memory as bounding boxes with some padding to include additional context, match the goal only against instances of the same object category, match the goal with the instance with the maximum matching score across all views. Further details are in Section~\ref{sec:offline} in the Supplementary Materials.

\begin{figure}[h]
\centering
\includegraphics[width=\textwidth]{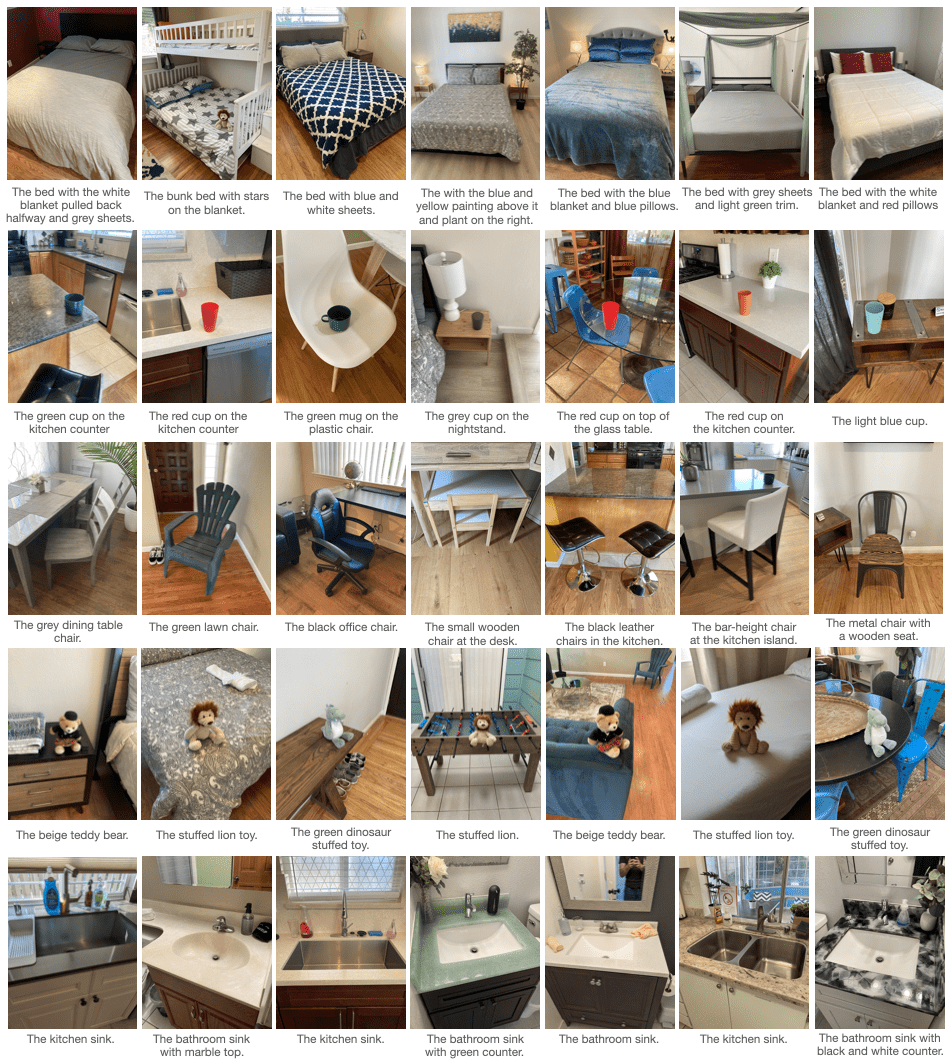}
\caption{
\textbf{Example goal object instances.}
Image and language descriptions used to uniquely identify target object instances.
}
\label{fig:object_instances1}
\end{figure}

\begin{figure}[h]
\centering
\includegraphics[width=\textwidth]{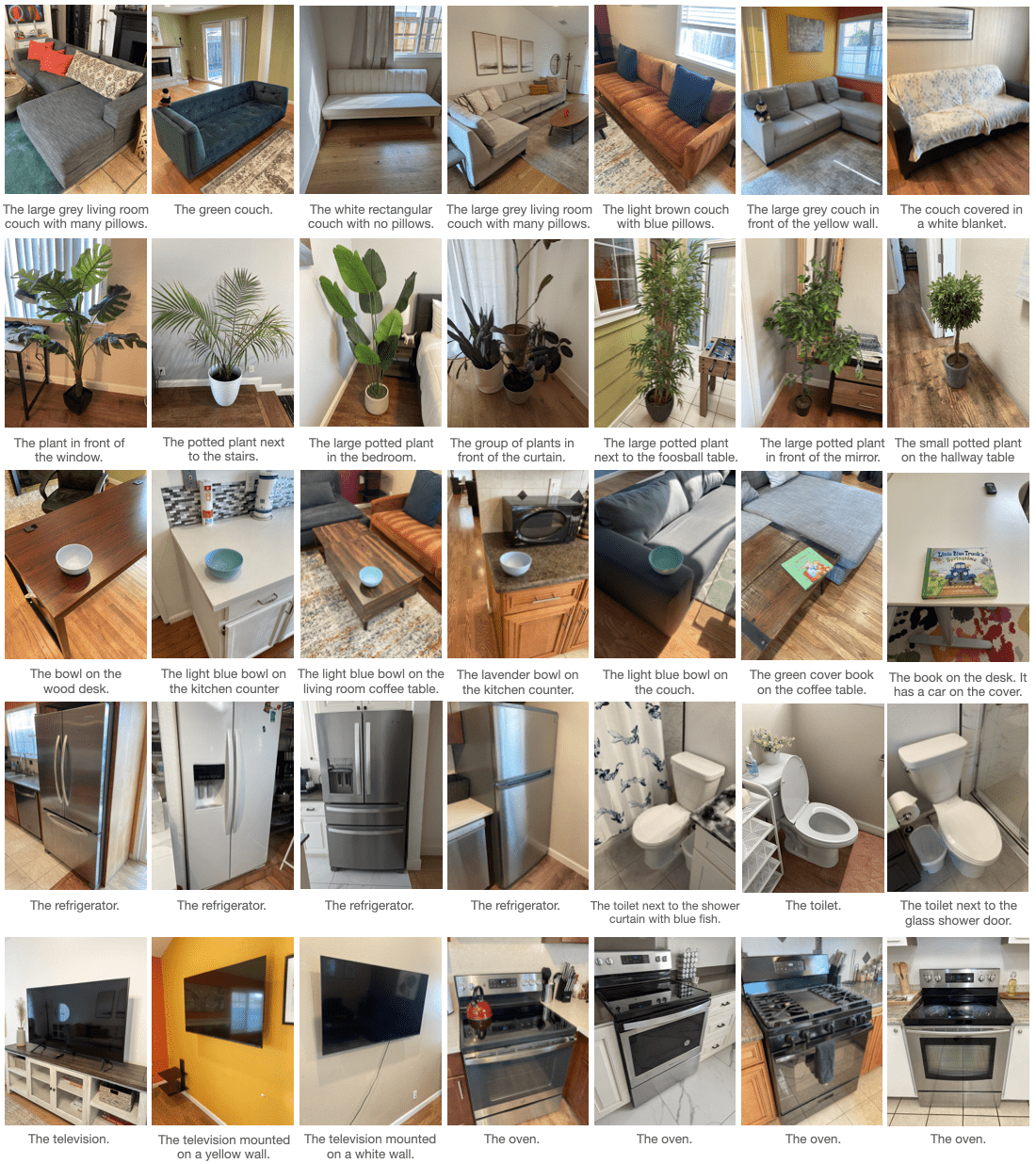}
\caption{
\textbf{More example goal object instances.}
Image and language descriptions used to uniquely identify target object instances.
}
\label{fig:object_instances2}
\end{figure}

\paragraph{Experimental Setting} We evaluate the GOAT agent as well as three baselines in nine visually diverse homes (see Figure \ref{fig:environment_diversity}) with 10 episodes per home consisting of 5-10 object instances randomly selected out of objects available in the home, representing 200+ different object instances in total (see Figures \ref{fig:object_instances1} and \ref{fig:object_instances2}). We selected goals across 15 different object categories (`chair', `couch', `potted plant', `bed', `toilet', `tv', `dining table', `oven', `sink', `refrigerator', `book', `vase', `cup', `bottle', `teddy bear'), took a picture for image goals following the protocol in Krantz~\textit{et al.}~\cite{krantz2023navigating}, and annotated 3 different language descriptions uniquely identifying the object. To generate an episode within a home, we sampled a random sequence of 5-10 goals split equally among language, image, and category goals among all object instances available. We evaluate approaches in terms of success rate to reach the goal and SPL~\cite{anderson2018evaluation}, which measures path efficiency as the ratio of the agent's path length over the optimal path length. We report evaluation metrics per goal within an episode with two standard deviation error bars.

\paragraph{Baselines} We compare GOAT to three baselines: \textbf{1.} \textbf{CLIP on Wheels}~\cite{gadre2023cows} - the existing work that comes closest to being able to address the GOAT problem setting - which keeps track of all images the robot has ever seen and, when given a new goal object, decides whether the robot has already seen it by matching CLIP~\cite{radford2021learning} features of the goal image or language description with CLIP features of all images in memory, \textbf{2.} \textbf{GOAT w/o Instances}, an ablation that treats all goals as object categories, i.e., always navigating to the closest object of the correct category instead of distinguishing between different instances of the same category as in \cite{gervet2023navigating}, allowing us to quantify the benefits of GOAT's instance awareness, and \textbf{3.} \textbf{GOAT w/o Memory}, an ablation that resets the semantic map and Object Instance Memory after every goal, allowing us to quantify the benefits of GOAT's lifelong memory.

\begin{table}[t!]
\caption{\textbf{Navigation Performance in Unseen Natural Home Environments.} 
We compare GOAT to three baselines in 9 unseen homes with 10 episodes per home consisting of 5-10 image, language, or category goal object instances in terms of success rate and SPL~\cite{anderson2018evaluation}, a measure of path efficiency, per goal instance. 
}
\label{tab:online_results}
\begin{adjustbox}{width=\textwidth}
\begin{tabular}{r|cccc|cccc}
\toprule
\multicolumn{1}{l|}{}        & \multicolumn{4}{c|}{SR per Goal}       & \multicolumn{4}{c}{SPL Per Goal}       \\
\multicolumn{1}{l|}{}        & Image & Language & Category & Average & Image & Language & Category & Average \\ \midrule
GOAT                         & $\mathbf{86.4 \pm 1.1}$  & $\mathbf{68.2 \pm 1.5}$     &  $\mathbf{94.3 \pm 0.8}$     & $\mathbf{83.0 \pm 0.7}$    & $\mathbf{0.679 \pm 0.013}$  & $\mathbf{0.511 \pm 0.014}$     & $\mathbf{0.737 \pm 0.010}$     & $\mathbf{0.642 \pm 0.007}$    \\
CLIP on Wheels               & $46.1 \pm 1.8$    & $40.8 \pm 1.9$       & $65.3 \pm 1.5$       & $50.7 \pm 1.0$       & $0.368 \pm 0.014$    & $0.317 \pm 0.013$       & $0.569 \pm 0.015$       & $0.418 \pm 0.008$      \\
GOAT w/o Instances & $28.6 \pm 1.7$  & $27.6 \pm 1.6$     & $\mathbf{94.1 \pm 0.8}$     & $49.4 \pm 0.8$    & $0.219 \pm 0.013$  & $0.222 \pm 0.012$     & $\mathbf{0.739 \pm 0.011}$     & $0.398 \pm 0.007$     \\
GOAT w/o Memory         & $59.4 \pm 1.5$  & $45.3 \pm 1.6$     & $76.4 \pm 1.3$     &  $60.3 \pm 0.8$   & $0.193 \pm 0.020$  & $0.134 \pm 0.022$     & $0.239 \pm 0.021$     & $0.188 \pm 0.012$    \\ \bottomrule
\end{tabular}
\end{adjustbox}
\end{table}

\begin{figure}[t!]
    \centering
    \includegraphics[width=\textwidth,keepaspectratio]{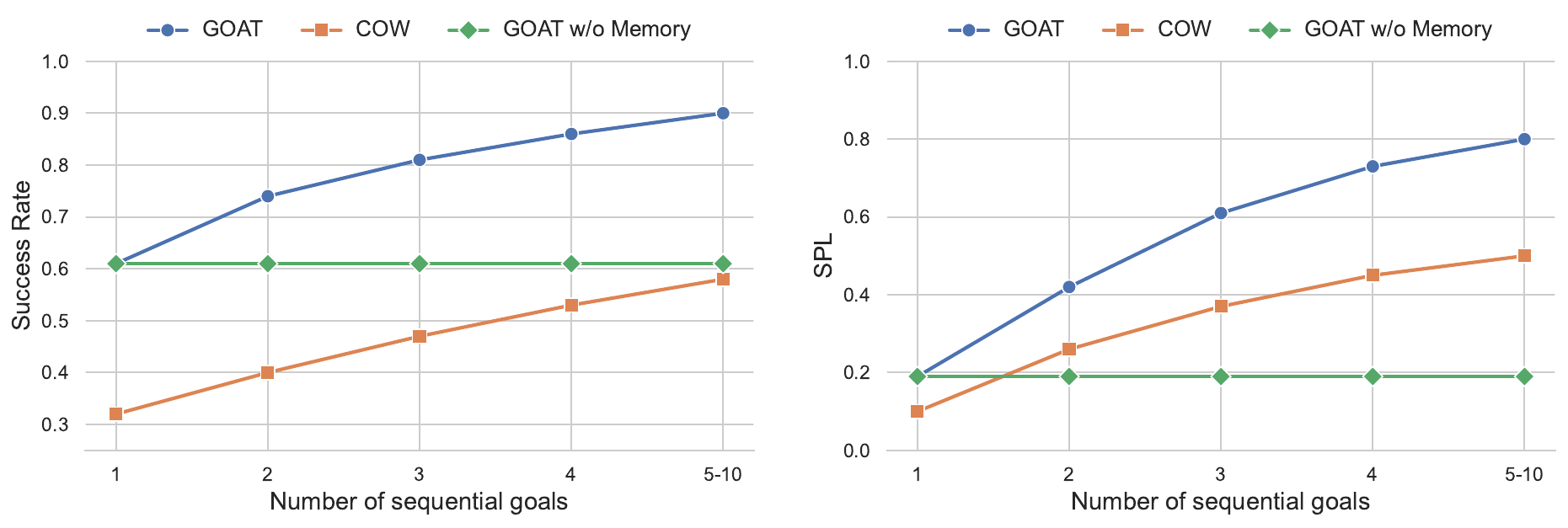}
    \caption{
    \small
    \textbf{Navigation performance based on sequential goal count.} GOAT performance improves with experience in the environment: from a 60\% success rate (0.2 SPL) at the first goal to 90\% (0.8 SPL) for goals 5-10 after thorough exploration. Conversely, GOAT without memory shows no improvement from experience, while COW benefits but plateaus at much lower performance.}
    \label{fig:goat_increasing_performance}
\end{figure}

\begin{figure}[!]
    \centering
    \thispagestyle{empty}
    \includegraphics[width=0.88\textwidth,keepaspectratio]{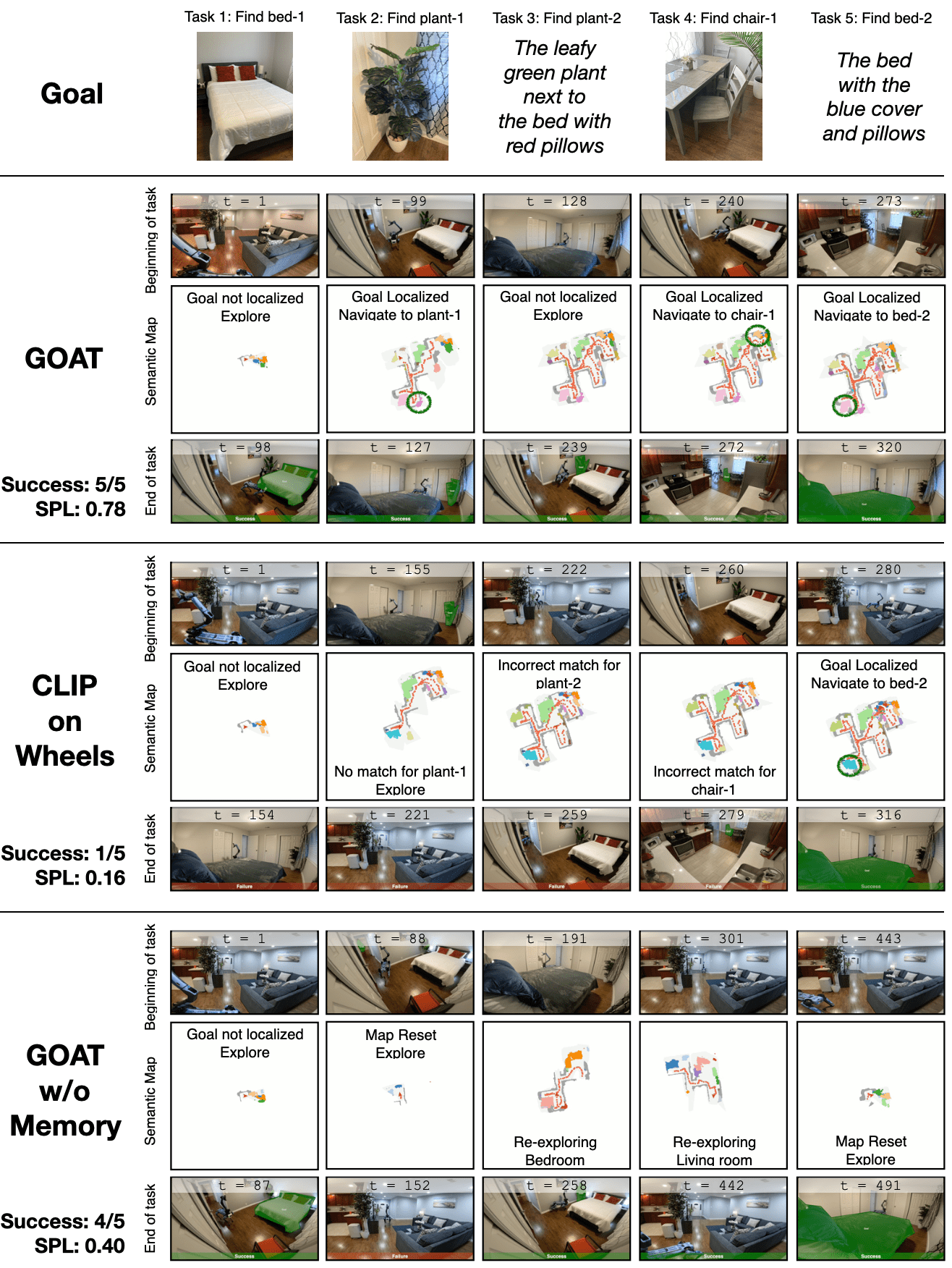}
    \caption{
    \small
    \textbf{Online evaluation qualitative trajectories.} We compare methods on the same sequence of 5 goals (top) in the same environment. GOAT localizes all goals and navigates efficiently (with an SPL of 0.78). CLIP on Wheels localizes only 1 out of 5 goals, illustrating the superiority of GOAT’s Object Instance Memory for matching. GOAT without memory is able to localize 4 our of 5 goals, but with an SPL of only 0.40 as it has to re-explore the environment with every goal. See Section \ref{sec:online} for details.}
    \label{fig:navigation_trajectory}
\end{figure}

\paragraph{Quantitative Results} Table \ref{tab:online_results} reports metrics for each method aggregated over the 90 episodes. GOAT achieves 83\% average success rate (94\% for object categories, 86\% for image goals, and 68\% for language goals). 
We observed that localizing language goals is harder than image goals (detailed in the Discussions section).
CLIP on Wheels~\cite{gadre2023cows} attains a 51\% success rate, showing that using GOAT's Object Instance Memory for goal matching is more effective than CLIP feature matching against all previously viewed images. GOAT w/o Instances achieves 49\% success rate, with 29\% and 28\% success rates for image and language goals respectively. This demonstrates the need to keep track of enough information in memory to be able to distinguish between different object instances, which~\cite{gervet2023navigating} wasn't able to do. GOAT w/o memory achieves 61\% success rate with an SPL of only 0.19 compared to the 0.64 of GOAT. It has to re-explore the environment with every goal, explaining the low SPL and low success rate due to many time-outs. This demonstrates the need to keep track of a lifelong memory. Figure~\ref{fig:goat_increasing_performance} further emphasizes this point: GOAT performance improves with experience in the environment from a 60\% success rate (0.20 SPL) at the first goal to 90\% (0.80 SPL) for goals 5-10 after thorough exploration. Conversely, GOAT without memory shows no improvement from experience, while COW benefits but plateaus at much lower performance.

\paragraph{Qualitative Results} We visualize representative trajectories in Figure \ref{fig:navigation_trajectory}. Here we show the performance of GOAT, CLIP on wheels, and GOAT without Memory on the same sequence of 5 goals, from the same initialization point. 
When matching image or language goals CLIP on Wheels computes features of the entire observed frame. This makes the matching threshold hard to tune, leading to more false positives (matches the wrong bed for task 1) and false negatives (misses the correct plant for task 2, eventually matching the incorrect plant). 
Without memory, the GOAT agent will continue to re-explore previously seen regions (tasks 3 and 4 re-explore previously explored already). 
Additionally, matching performance is worse because the agent forgets previously observed instances. Matching performance improves as more of the environment is explored and mapped because the effect of the matching threshold is diminished (see Section \ref{sec:offline} for details).
The full GOAT system can handle these issues. GOAT is able to match all instances and efficiently navigate to them correctly.

\subsection{Applications}

\begin{figure}[!]
    \centering
    \includegraphics[width=0.88\textwidth,keepaspectratio]{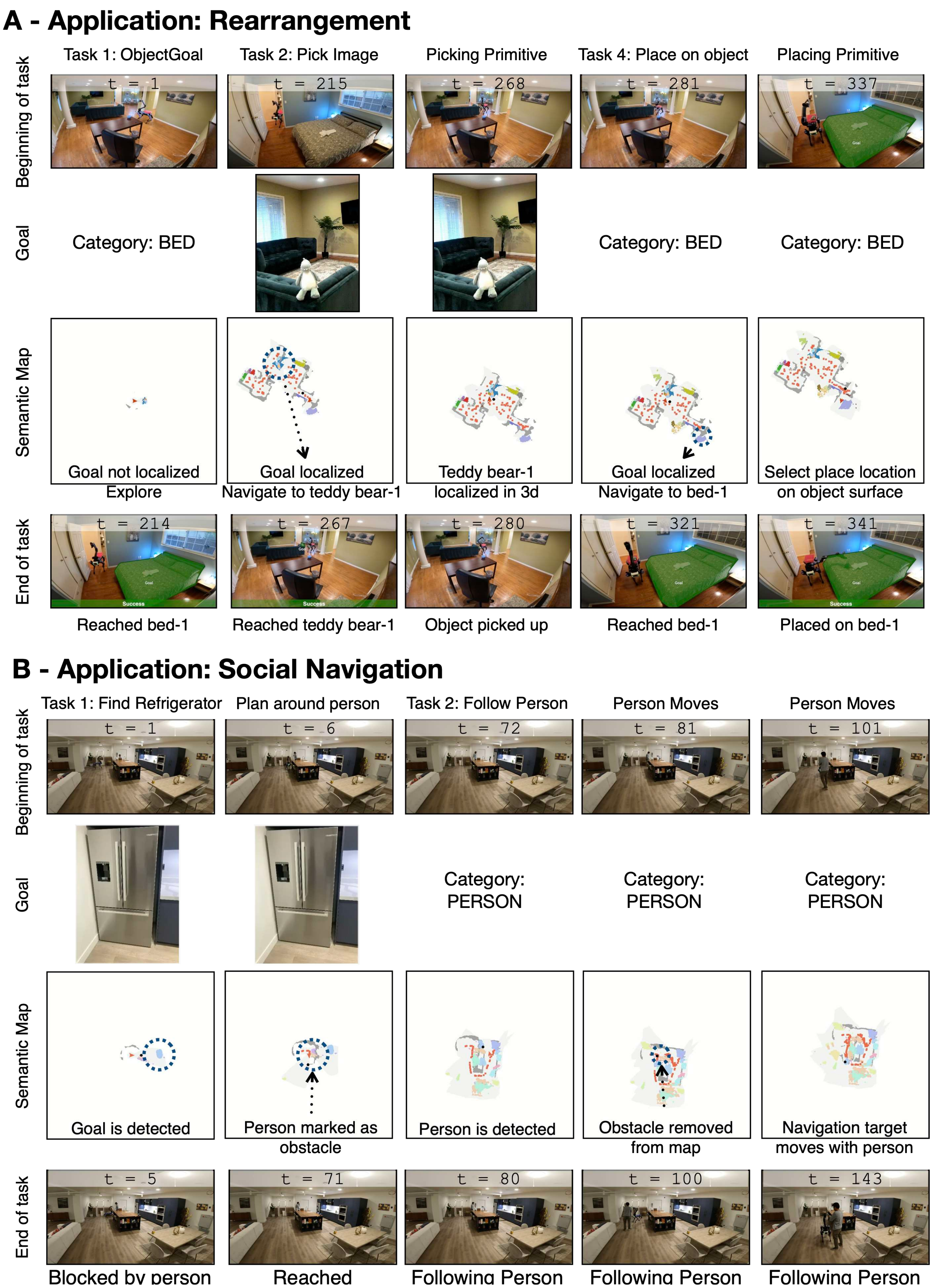}
    \caption{
    \small
   \textbf{A - Application: Rearrangement.} The GOAT policy searches for then picks up a toy and places it on the bed.
   \textbf{B - Application: Social Navigation.} The GOAT policy finds a refrigerator while avoiding a person, then follows a person.
    }
    \label{fig:pick_place}
\end{figure}

As a general navigation primitive, the GOAT policy can readily be applied to downstream tasks such as pick and place and social navigation.

\paragraph{Open Vocabulary Mobile Manipulation} The ability to perform rearrangement tasks is essential in any deployment scenarios for mobile robots (homes, warehouses, factories) \cite{batra2020rearrangement, ovmm, driess2023palm, pmlr-v205-ichter23a, gu2022multi}. These are commands such as ``pick up my coffee mug from the coffee table and bring it to the sink," requiring the agent to search for and navigate to an object, pick it up, search for and navigate to a receptacle, and place the object on the receptacle. The GOAT navigation policy can easily be combined with pick and place skills (we use built-in skills from Boston Dynamics) to fulfill such requests. We evaluate this ability on 30 such queries with image/language/category objects and receptacles across 3 different homes. GOAT can find objects and receptacles with 79\% and 87\% success rates, respectively. 

We visualize one such trajectory in Figure \ref{fig:pick_place} (A). The agent is tasked with first finding a bed, finding a specific toy, and then moving that toy to the bed. We see that while exploring for the bed, the agent observes the toy and keeps it in the instance memory. Consequently, after finding the bed, the agent is able to directly navigate back to the toy (column 2), then efficiently pick it up, and move it back to the bed (column 5).

\paragraph{Social Navigation} To operate in human environments, mobile robots need the ability to treat people as dynamic obstacles, plan around them, and search for and follow people~\cite{luber2012socially, puig2023habitat}. To give the GOAT policy such skills, we treat people as image object instances with the \emph{PERSON} category. This enables GOAT to deal with multiple people, just like it can deal with multiple instances of any object category. GOAT can then remove someone's previous location from the map after they have moved. To evaluate the ability to treat people as dynamic obstacles, we introduce moving people in 5 trajectories, otherwise following the same experimental setting as our main experiments. GOAT preserves an 81\% success rate. We further evaluate the ability of GOAT to search for and follow people by introducing such goals in 5 additional trajectories. GOAT can localize and follow people with 83\% success, close to the 86\% success rate for static image instance goals. 

We visualize a qualitative example of a trajectory in Figure \ref{fig:pick_place} (B). Here the agent must navigate to the refrigerator and then follow the human. We see that the agent recognizes the refrigerator (column 1), but the route there is blocked by the human so the agent must plan around (column 2). After reaching the refrigerator, the agent begins following the human while constantly updating the map based on new sensor observations. This allows the agent to move through space that had been previously marked as occupied by the person (column 4). The navigation target continues to track the person as they move around the apartment (column 5).

\section{Discussion}
\begin{figure}[!]
    \centering
    \adjustbox{width=1.1\textwidth, center=\textwidth}{%
        \includegraphics{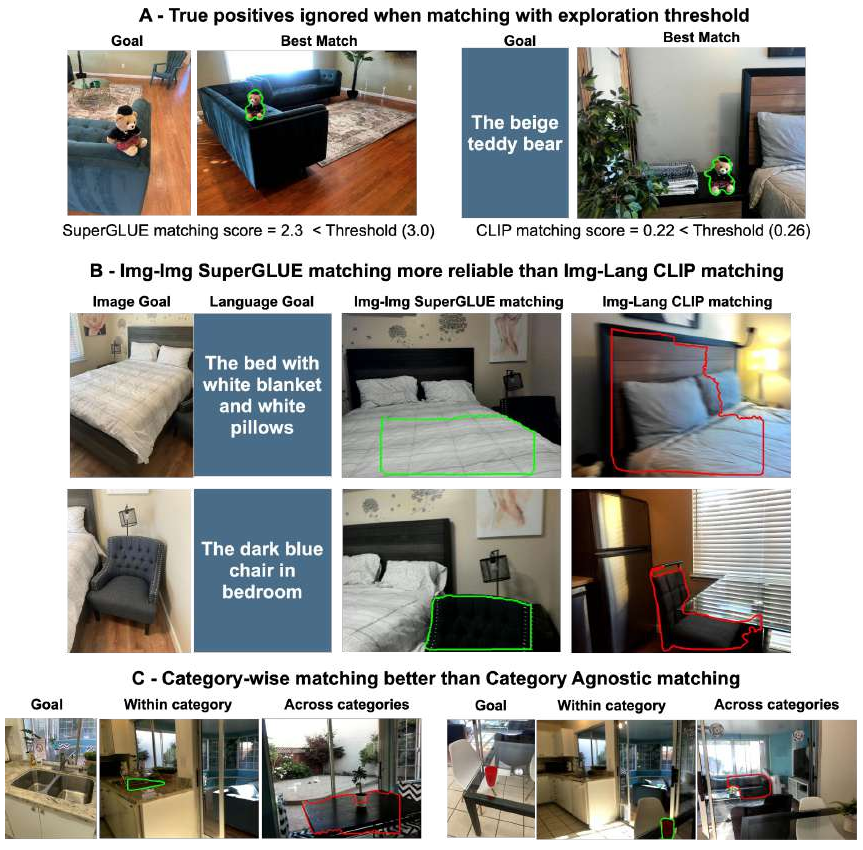}%
    }
    \caption{\textbf{Qualitative examples of trends observed in matching.} 
    (A) Matching with a threshold during exploration can result in false negatives, which would be correct matches post-exploration. 
    (B) Image-image SuperGLUE is more reliable than image-language CLIP matching. 
    (C) Matching within a category performs better than matching across categories. 
    }
    \label{fig:trends}
\end{figure}

\paragraph{Modularity allows GOAT to Achieve Robust General-Purpose Navigation in the Real World}

The GOAT system as a whole is a robust navigation platform, achieving a success rate of 83\% across image, language, and category goals in the wild (up to 90\% once the environment is fully explored). This is possible in-part due to the modular nature of the system. A modular system allows learning to be applied in the components in which it is required (\ie object detection, image/language matching), while still leveraging strong classical methods (\ie mapping and planning). Furthermore, for learning-based components, we can use models trained on large datasets (\ie CLIP, MaskRCNN), or specialized tasks (monocular depth estimation) to full effect, where a task-specific end-to-end learned approach would be limited by the available data for this specific task. GOAT is able to tie all of these components together using our Object Instance memory to achieve state-of-the-art performance for lifelong real-world navigation. 

Furthermore, the modular design of GOAT allows it to be easily adapted to different robot embodiments and a variety of downstream applications. 
GOAT can be deployed on any robot with an RGB-D camera, a pose sensor (onboard SLAM), and the ability to execute low-level locomotion commands (move forward, turn left, turn right).
GOAT's modularity eliminates the need for new data collection or training when deployed on a new robot platform. This stands in contrast to end-to-end methods, which would require new data collection and retraining for every different embodiment. 

\paragraph{Matching Performance During Exploration Lags Behind Performance After Exploration}
Using a predefined threshold for a successful goal to object matching score during exploration (on the fly) as goals is tricky because an inflexible threshold can cause true positives to be ignored (Figure~\ref{fig:trends}-A) and false positives to be counted in. On the other hand, once the scene has been explored, the agent has the privilege of selecting the instance with the best matching score as the goal. This is reflected in improved performance of the agent post-exploration (6\% higher success rate). Refer to Section~\ref{sec:offline} Table~\ref{tab:offline_matching_ablations} in the Supplementary Materials for details.

\paragraph{Image Goal Matching is More Reliable than Language Goal Matching}

We observe that image-to-image goal matching is more successful at identifying goal instances as compared to matching instance views with semantic features of language descriptions of the goal.
This is expected because SuperGLUE-based image keypoint matching can leverage correspondences in geometric properties between predicted instances and goal objects. However, the semantic feature encodings from CLIP can be incapable of capturing fine-grained instance properties – that can often be crucial for goal matching (see examples in Figure~\ref{fig:trends}-B). As a result, matching instance views with image goals is 23\% more successful than matching with language description features.

\paragraph{Goal Matching Improves by Subsampling Instances by Category and Adding Context}

When sifting through seen instances to find a match with the goal, the agent can either compare against all instances seen so far, or do this comparison only against instances that belong to the goal category. We observe that filtering out non-goal categories improves matching accuracy by 23\% – preventing false positives from getting matched (Figure~\ref{fig:trends}-C). Moreover, this is computationally also better – as comparing only against a subset of instances is also faster and more efficient. Additionally, regardless of whether we use SuperGLUE or CLIP for matching instances to goals, we observe that providing more context about the instance's background – using wider, enlarged bounding boxes – results in improved matching accuracy (up to 22\% more successful than matching bounding boxes alone).

\paragraph{Real-World Open-Vocabulary Detection: Limitations and Opportunities}

An interesting and noteworthy observation is that despite the rapid advances in open (or large) vocabulary vision-and-language models (VLMs)~\cite{lseg, owlvit}, we find their performance to be significantly worse than a Mask RCNN model from 2017. We attribute this observation to two possible hypotheses: (i) open-vocabulary models trade-off robustness for being more versatile, and supporting more queries, and (ii) the internet-scale weakly labeled data sources used to train modern VLMs under-represent the kind of embodied interaction data that would benefit robots occupying real-world environments with humans. The latter represents a challenging opportunity to develop such large-scale models that are simultaneously versatile and robust for embodied applications in real-world environments.

\section{Materials and Methods}
\subsection{Go To Any Thing System Architecture}

\paragraph{System Overview} Figure \ref{fig:goat_architecture1.png} (A) shows an overview of the GOAT system. The perception system detects object instances, localizes them in a top-down semantic map of the scene, and stores images in which each instance has been viewed in an Object Instance Memory. When a new goal is specified, a global policy first tries to localize it within the Object Instance Memory. If no instance is localized, the global policy outputs an exploration goal. A local policy finally computes actions towards the long-term goal.

\begin{figure}
    \centering
    \includegraphics[width=\textwidth]{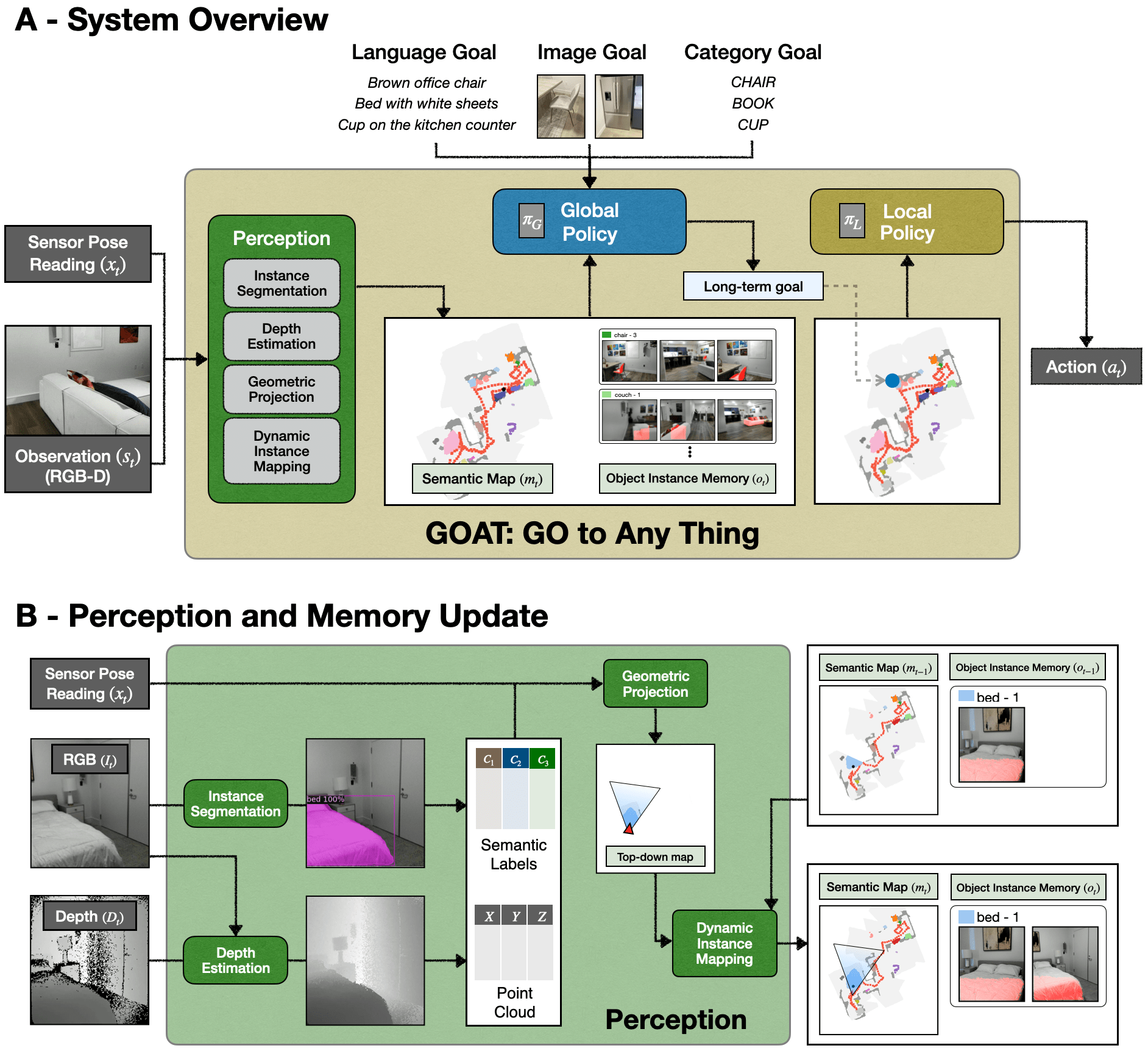}
    \caption{
    \small
    \textbf{(A) GOAT system overview.} The perception system detects and localizes object instances, the global policy outputs high-level navigation commands depending on whether the robot should explore or reach a goal already in memory, and the local policy executes these commands. \textbf{(B) Perception and memory update.} The perception system processes RGB-D input to infill depth, segment object instances, project them into a top-down semantic map, and store views in the Object Instance Memory.
    }
    \label{fig:goat_architecture1}
\end{figure}

\begin{figure}
    \centering
    \includegraphics[width=\textwidth]{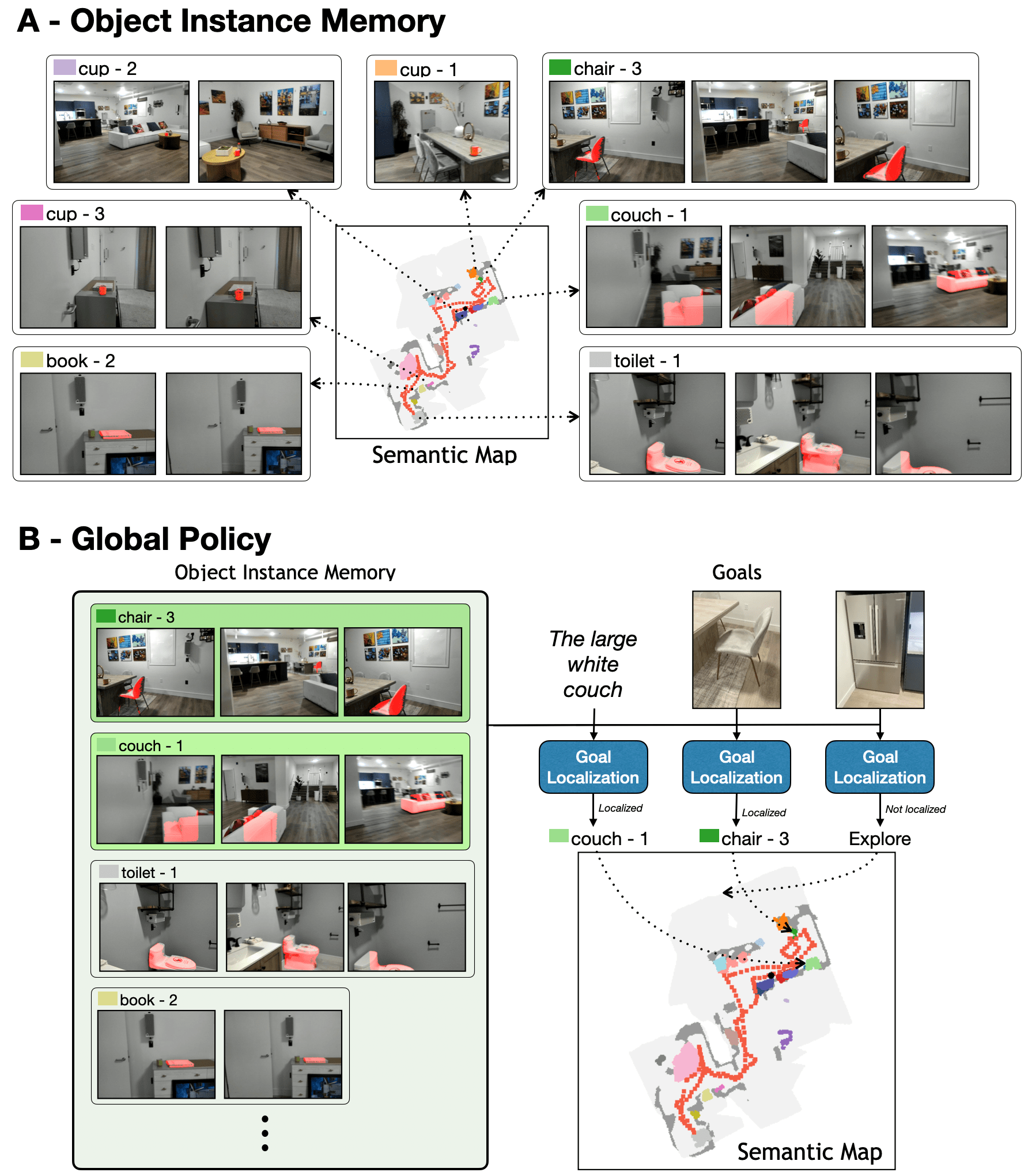}
    \caption{
    \small
    \textbf{(A) Object Instance Memory.}
    We cluster object detections, along with image views in which they were observed, into instances using their location in the semantic map and their category.
    \textbf{(B) Global Policy.}
    When a new goal is specified, the global policy first tries to localize it within the Object Instance Memory. If no instance is localized, it outputs an exploration goal.
    }
    \label{fig:goat_architecture2}
\end{figure}

\paragraph{Perception:} Figure \ref{fig:goat_architecture1} (B) shows perception system. It takes as input the current depth image $D_t$, RGB image $I_t$, and pose reading $x_t$ from onboard sensors. It uses an off-the-shelf model to segment instances in the RGB image. We use MaskRCNN~\cite{he2017mask} with a ResNet50~\cite{he2016deep} backbone pretrained on MS-COCO for object detection and instance segmentation. We chose MaskRCNN as current open-set models, such as Detic~\cite{zhou2022detecting}, were less reliable for common categories in early experiments. We also estimate depth to fill in holes for reflective objects in raw sensor readings. 

To fill holes in the depth image we first use a monocular depth estimation model to obtain a dense depth estimation from $I_t$ (we used the MiDaS~\cite{midas} model, although any depth estimation model would be applicable). Since depth estimation models typically predict relative distances instead of absolute distances, we ground the predicted depth using the known true depth values from $D_t$. Specifically, we solve for the scale factor that minimizes the mean squared error between estimated depth and sensed depth across all pixels where there is a depth reading.
\[\argmin_{A,b} \sum_i \|D_{t,i}-AX_{t,i}-b\|_2 \]
Where $D_{t,i}$ is the $i$th depth reading from the depth reading $D_t$ and $X_{t,i}$ is the depth estimate at the same point. This optimization can be easily solved as a system of equations yielding a scale factor and offset to project estimated depth points into absolute distances. We use these depth estimates for pixels in the depth image for which no reading was registered.

Using the dense depth computed above, we project the first-person semantic segmentation into a point cloud, bin the point cloud into a 3D semantic voxel map, and finally sum over the height to compute a 2D instance map $m_t$. For each detected object instance, we also store the image in which the object was detected as part of the object instance memory.

\paragraph{Semantic Map Representation:} 
The semantic map ($m_t$ at timestep $t$) is a spatial representation of the environment that keeps track of object instance locations, obstacles, and explored areas. Concretely, it is a $K \times M \times M$ matrix of integers where $M \times M$ is the map size, and $K$ is the number of map channels. Each cell of this spatial map corresponds to $25 \text{cm}^2$ ($5 \text{cm} \times 5 \text{cm}$) in the physical world. 
Map channels $K = C + 4$ where $C$ is the number of semantic object categories, and the remaining $4$ channels represent the obstacles, the explored area, and the agent’s current and past locations. 
An entry in the map is non-zero if the cell contains an object of a particular semantic category, an obstacle, or is explored, depending on the channel, and zero otherwise. 
In this semantic map representation, the first $C$ channels store the unique instance ids of the projected objects.
The map is initialized with all zeros at the beginning of an episode, and the agent starts at the center of the map facing east.

\paragraph{Object Instance Memory} Figure \ref{fig:goat_architecture2} (A) shows the Object Instance Memory ($o_t$ at timestep $t$). Our object instance memory clusters object detections across time into instances using their location in the map and their category. 

Each object instance $i$ is stored as a set of cells in the map $C_i$, a set of object views represented as bounding boxes with context $M_i$, and an integer indicating the semantic category $S_i$. 
For each incoming RGB image $I$, we detect objects. For each detection $d$ we use the bounding box around the detection $I_d$, the semantic class $S_d$, and the corresponding points in the map $C_d$ based on projected depth. We dilate each instance on the map $C_d$ by $p$ units to obtain a dilated set of points 
  $D_d$ per instance, which is used for matching to instances that were previously added in the memory and map.
We check for matches pairwise between each detection and each existing object instance. A detection $d$ and instance $i$ are considered to match if the semantic category is the same, and any projected locations in the map overlap, $S_d = S_i$ and $D_d \cap C_i \ne 	\varnothing $. If there is a match, we update the existing instance with the new points and new image
\[C_i \leftarrow C_i \cup C_d, M_i \leftarrow M_i \cup \{I_d\}\] 
Otherwise, a new instance is added using $C_d$ and $I_d$. 

This procedure aggregates unique object instances over time, allowing new goals to be matched against all images of specific instances or categories easily. 

\paragraph{Global Policy} Figure \ref{fig:goat_architecture2} (B) shows the global policy. When a new goal is specified to the agent, the global policy $\pi_G$ first searches the Object Instance Memory to see if the goal has already been observed. The method for computing matches is tailored to the modality of the goal specification. For category goals, we simply check whether any object of the category is in the semantic map. For language goals, we first extract an object category from the language description (by prompting with Mistral 7B~\cite{jiang2023mistral} in our experiments), then match CLIP features of the language description with CLIP features of each object instance of the inferred category in our Object Instance Memory. Similarly, for image goals, we first extract an object category from the image with MaskRCNN, then match keypoints of the goal image with keypoints of each object instance of the inferred category with an off-the-shelf SuperGlue model. While the environment is being explored, we consider the object instance matches the goal if the matching score is above some threshold ($0.28$ for CLIP, $6.0$ for Superglue), while when the environment is fully explored, we select the object instance with the highest similarity score. After an instance is selected, its stored location in the top-down map is used as a long-term point navigation goal. If no instance is localized, the global policy outputs an exploration goal. We use frontier-based exploration~\cite{yamauchi1998frontier}, which selects the closest unexplored region as the goal.

\paragraph{Local Policy} Given a long-term goal output by the global policy $\pi_G$, the local policy $\pi_L$ uses the Fast Marching Method to plan a path towards it. On the Spot robot, we use the built-in point navigation controller to reach waypoints along this path. On the Stretch robot with no such built-in controller, we plan the first low-level action along this path deterministically as in~\cite{gervet2023navigating}. 

\subsection{Experimental Methodology}

\paragraph{Hardware Platforms} The GOAT navigation policy is platform agnostic: no component of our system is tied to any particular robot hardware. We deployed GOAT on and conducted qualitative experiments with Boston Dynamics Spot and Hello Robot Stretch robots. We conducted large-scale quantitative experiments with GOAT on Spot (due to its higher reliability) against 3 baselines in 9 real-world homes to reach a total of 200+ different object instances (see Figure ~\ref{fig:environment_diversity}).

\paragraph{Navigation Performance in Unseen Natural Home Environments} We evaluate GOAT “in the wild” in 9 unseen rented homes without pre-computed maps or locations of objects. We evaluate each method for 10 trajectories per home with 5-10 goals per trajectory for a total of 90 hours of experiments. We selected goals across 15 different object categories ("chair", "couch", "potted plant", "bed", "toilet", "tv", "dining table", "oven", "sink", "refrigerator", "book", "vase", "cup", "bottle", "teddy bear"), took a picture for image goals following the protocol in ~\cite{krantz2023navigating}, and annotated 3 different language descriptions uniquely identifying the object. To generate an episode within a home, we sampled a random sequence of 5-10 goals split equally among language, image, and category goals among all object instances available.

\paragraph{Metrics} We report metrics per goal within an episode, as compound metrics over the entire trajectory are not meaningful. We consider navigation to a goal within an episode successful if the robot called the stop action close enough (less than 1 meter) to the correct instance of the goal category within a reasonable time budget ($200$ robot steps). To compute the Success weighted by Path Length (SPL)~\cite{anderson2018evaluation} per goal, we measure the geodesic distance to the goal instance closest to the previous goal instance. 

\clearpage
\section{Supplementary}
\subsection{Offline Comparison of Instance Matching Strategies}
\label{sec:offline}
In this section, we compare design choices for the matching module of the global policy, whose role is to identify the goal object instance among previously seen object instances. This module is particularly important as it determines the form of the Object Instance Memory and allows the GOAT agent to perform lifelong learning for multimodal navigation. Recall that our matching module uses CLIP for matching language goals and SuperGLUE for matching image goals. We first filter instances by target category and use a cropped version of each instance view by including some context around the object. We then aggregate the scores across views via a "max" operation. During exploration, we use a threshold of 3.0 for image-image SuperGLUE matching and a threshold of 0.75 for language-image CLIP matching. Post-exploration, we pick the best matching instance without using any threshold.

\begin{table}[!]
    \centering
\begin{tabular}{|l|l|c|c|c|cccc|}\hline
       & & Threshold  & Sub-sampling  & Context & {\cellcolor{green} Max} & Median & Avg & Avg Top-2  \\
    \hline
  \multirow{24}{*}{\rotatebox[origin=c]{90}{Image-to-image}} & \multirow{12}{*}{\rotatebox[origin=c]{90}{SuperGLUE}} & \multirow{6}*{\begin{tabular}{c} 
    Exploration\\
    threshold\\
    = 3.0\\
\end{tabular}} &\multirow{3}{*}{All images} & Bbox & 48.4 & 25.8 & 41.9 & 46.8\\ \cline{5-5}
& & && Bbox + pad & 72.6 & 56.5 & 62.9 & 67.7\\ \cline{5-5}
& & && Full image & 58.1 & 40.3 & 40.3 & 53.2\\ \cline{4-5}

& & &\multirow{3}{*}{By category} & Bbox & 51.6 & 25.8 & 43.5 & 48.4\\ \cline{5-5}
& & && {\cellcolor{green} Bbox + pad} & {\cellcolor{lime}\textbf{92.1}} & 83.0 & 85.1 & 89.4\\ \cline{5-5}
& & && Full image & 91.9 & 80.6 & 85.5 & 90.3\\  \cline{3-9}
  
  && \multirow{6}*{\begin{tabular}{c} 
    No\\
    threshold\\
\end{tabular}} & \multirow{3}{*}{All images} & Bbox & 50.0 & 38.7 & 50.0 & 51.6\\ \cline{5-5}
& & && Bbox + pad & 72.6 & 58.1 & 64.5 & 67.7\\ \cline{5-5}
& & && Full image & 58.1 & 40.3 & 41.9 & 53.2\\ \cline{4-5}

& & &\multirow{3}{*}{By category} & Bbox & 64.5 & 43.5 & 64.5 & 62.9\\ \cline{5-5}
& & && {\cellcolor{green} Bbox + pad} & {\cellcolor{lime}\textbf{95.2}} & 85.1 & 89.4 & 93.6\\ \cline{5-5}
& & && Full image & 91.9 & 85.5 & 88.7 & 91.9\\  \cline{2-9}

  & \multirow{12}{*}{\rotatebox[origin=c]{90}{CLIP}} & \multirow{6}*{\begin{tabular}{c} 
    Exploration\\
    threshold\\
    = 0.75 \\
\end{tabular}} & \multirow{3}{*}{All images} & Bbox & 40.3 & 17.7 & 25.8 & 38.7\\ \cline{5-5}
& & && Bbox + pad & 48.4 & 33.9 & 29.0 & 45.2\\ \cline{5-5}
& & && Full image & 46.8 & 30.6 & 32.3 & 50.0\\ \cline{4-5}

& & &\multirow{3}{*}{By category} & Bbox & 53.2 & 22.6 & 25.8 & 46.8\\ \cline{5-5}
& & && {\cellcolor{green} Bbox + pad} & {\cellcolor{lime}\textbf{79.0}} & 64.5 & 72.6 & 74.2\\ \cline{5-5}
& & && Full image & {\cellcolor{lime} \textbf{79.0}} & 56.5 & 59.7 & 74.2\\  \cline{3-9}
  
  && \multirow{6}*{\begin{tabular}{c} 
    No\\
    threshold\\
\end{tabular}} & \multirow{3}{*}{All images} & Bbox & 40.3 & 22.6 & 29.0 & 38.7\\ \cline{5-5}
& & && Bbox + pad & 48.4 & 35.5 & 30.6 & 46.8\\ \cline{5-5}
& & && Full image & 46.8 & 32.3 & 33.9 & 50.0\\ \cline{4-5}

& & &\multirow{3}{*}{By category} & Bbox & 66.1 & 56.5 & 61.3 & 66.1\\ \cline{5-5}
& & && {\cellcolor{green} Bbox + pad} & {\cellcolor{lime}\textbf{82.5}} & 67.7 & 75.8 & 75.8\\ \cline{5-5}
& & && Full image & 82.3 & 71.0 & 72.6 & 79.0\\ \cline{1-9}
    \hline
  \multirow{12}{*}{\rotatebox[origin=c]{90}{Language-to-image}} & \multirow{12}{*}{\rotatebox[origin=c]{90}{CLIP}} & \multirow{6}*{\begin{tabular}{c} 
    Exploration\\
    threshold\\
    = 0.26 \\
\end{tabular}} &\multirow{3}{*}{All images} & Bbox & 54.8 & 33.9 & 25.8 & 53.2\\ \cline{5-5}
& & && Bbox + pad & 48.4 & 24.2 & 22.6 & 45.2\\ \cline{5-5}
& & && Full image & 38.7 & 17.7 & 22.6 & 38.7\\ \cline{4-5}

& & &\multirow{3}{*}{By category} & Bbox & 58.1 & 35.5 & 27.4 & 59.7\\ \cline{5-5}
& & && {\cellcolor{green} Bbox + pad} & {\cellcolor{lime} \textbf{69.4}} & 48.4 & 51.6 & 62.9\\ \cline{5-5}
& & && Full image & 59.7 & 48.4 & 45.2 & 53.2\\  \cline{3-9}
  
  && \multirow{6}*{\begin{tabular}{c} 
    No\\
    threshold\\
\end{tabular}} &\multirow{3}{*}{All images} & Bbox & 58.1 & 41.9 & 35.5 & 54.8\\ \cline{5-5}
& & && Bbox + pad & 51.6 & 30.6 & 29.0 & 50.0\\ \cline{5-5}
& & && Full image & 43.5 & 24.2 & 30.6 & 48.4\\ \cline{4-5}

& & &\multirow{3}{*}{By category} & Bbox & 71.0 & 62.9 & 58.1 &{\cellcolor{lime} \textbf{72.6}}\\ \cline{5-5}
& & && {\cellcolor{green} Bbox + pad} & {\cellcolor{lime}\textbf{72.6}} & 66.1 & 69.4 & 71.0\\ \cline{5-5}
& & && Full image & 69.4 & 67.7 & 67.7 & 66.1\\ \cline{1-9}
\end{tabular}
\caption{
\textbf{Offline comparison of instance matching strategies.}
We compare different design choices for the goal-to-instance matching module of the global policy on 27 trajectories manually annotated with ground-truth object instances corresponding to each goal. We report the matching success rate (higher the better), which is the percentage of goals that get correctly matched. The best entries for each matching method and matching threshold are highlighted in \colorbox{lime}{lime}, and the parameters used in online experiments are highlighted in \colorbox{green}{green}.
}
\label{tab:offline_matching_ablations}
\end{table}

We manually annotated 3 trajectories per home with ground-truth object instances corresponding to each goal, for a total of 27 trajectories. This enables us to evaluate the effect of different design choices on the matching success rate: the percentage of goals that get correctly matched. Table \ref{tab:offline_matching_ablations} presents ablations for the following design choices:

\begin{itemize}
    \item \textbf{Matching method}: Storing raw image views in our Object Instance Memory lets us use different matching methods per goal modality. We match language goal descriptions with object views in memory using the cosine similarity score between their CLIP~\cite{radford2021learning} features. On the other hand, to match image goals with object views in memory, we evaluate both CLIP feature matching and keypoint-based matching with SuperGLUE~\cite{sarlin2020superglue}.

    \item \textbf{Matching threshold}: The threshold for a successful matching score. We show results for a fixed non-zero threshold (the best hyper-parameter) and a zero threshold. We use the former when the agent is still exploring the scene because it has to decide whether the instance in the current observation matches the goal or to keep exploring, and the latter when the agent has already explored the entire scene and always navigates to the best match. Note that we assume that a match always exists. Hence when the agent has fully explored the environment, we expect the best match to be correct assuming the agent detects the object.
    
    \item \textbf{Instance sub-sampling}: Whether to compare the goal with views of all instances captured so far or only against instances of the goal's category. Intuitively, the latter is faster with higher precision but potentially lower recall, as it relies on accurate object detection.
    
    \item \textbf{Context}: The instance view context to use when matching: (i) only the bounding box crop of the detected instance (`bbox'), (ii) add some of the surrounding context (`bbox+pad'), or (iii) the full image in which the instance is seen (`full image').
    
    \item \textbf{Best match selection criteria}: When comparing multiple views of multiple instances with one goal, we can select the best match through: (i) max: choosing the instance with the maximum matching score (from any one view), (ii) median: the highest median matching score (across all views), (iii) the highest average matching score (across all views), and (iv) the highest average score across the top-k views.
    
\end{itemize}

Looking at the image-to-image matching section of Table \ref{tab:offline_matching_ablations}, we can see that:

\begin{enumerate}
    \item SuperGLUE-based image keypoint matching is much more reliable than CLIP feature matching — on average 13\% more successful. This helps explain the superior performance of GOAT over COW~\cite{gadre2023cows}, which uses CLIP feature matching.
    \item Introducing a matching threshold to ignore low confidence has a cost — on average 6\% worse than with no threshold. As we'll see in the Discussion, this means matching is more challenging during exploration than once the environment is fully explored.
    \item Sub-sampling instances based on goal category works better than sifting through all instances — on average 23\% more successful. This helps explain the superior performance of GOAT over COW, which doesn't subsample instances by category.
    \item Matching padded (enlarged) bounding box of instance views works best — on average 4.6\% better than using the full image (the second best approach) and 22\% better than using the object's bounding box across all settings. 
    \item Matching the maximum matching score across all views of all instances works better than median, average, and top-2 average — on average 2\% to 16\% better across all settings.
\end{enumerate}

Similar trends can be observed in language-to-image matching. However, image-to-image matching (using SuperGLUE-based keypoint matching) is much more reliable than (CLIP-based) language-to-image matching — on average 23\% better across all settings.

\begin{figure}
    \centering
    \includegraphics[width=\textwidth]{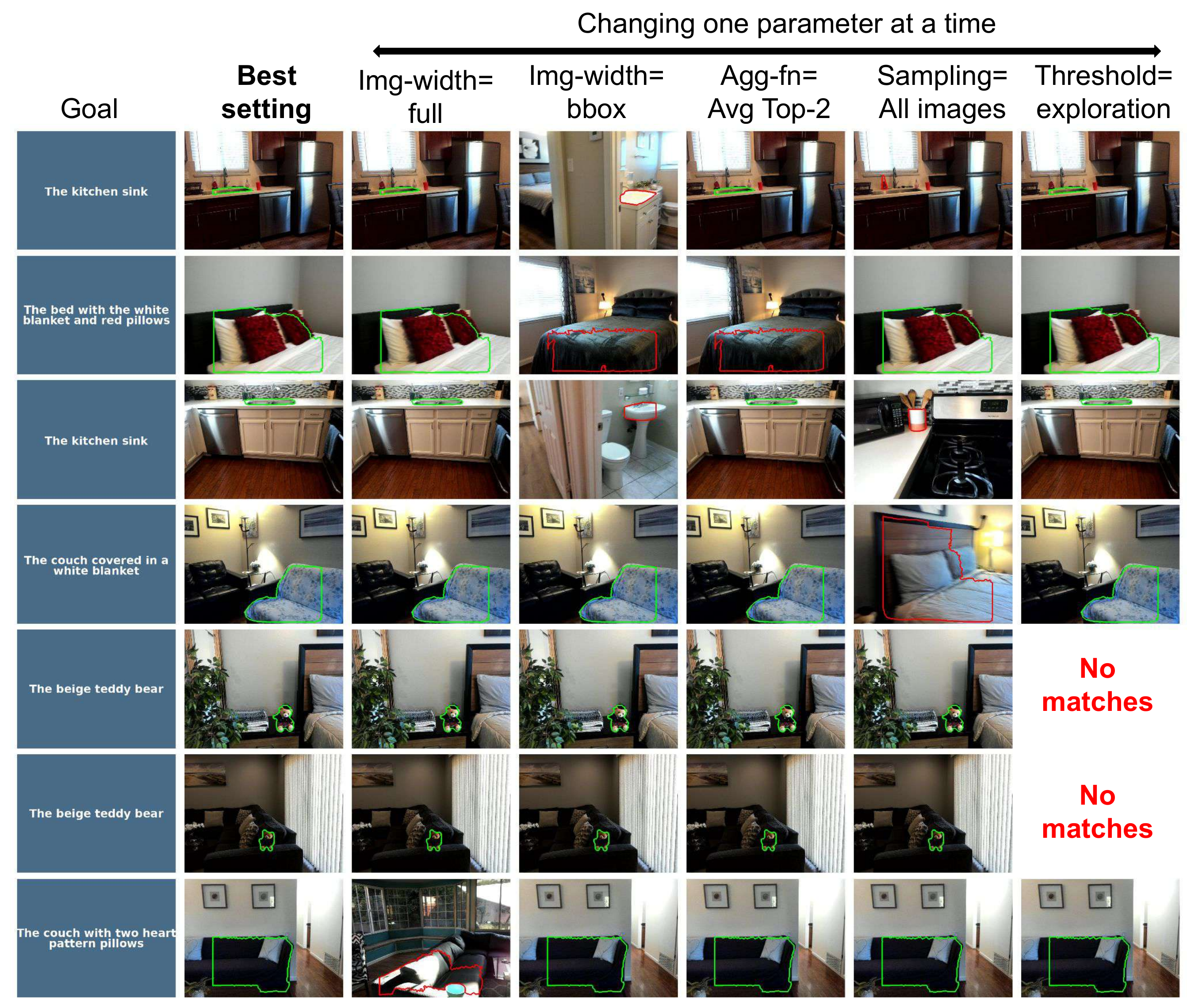}
    \caption{
    \textbf{Matching strategy ablation for selecting language goals.}
    In the second column, we show the language goals identified by our best setting (Img-width=Bbox+pad, Agg-fn=max, Sampling=By Category, Threshold=0.0). In the subsequent columns, we show results after changing one parameter at a time with respect to the best setting. The object instance selected in each setting is highlighted with a border which is green for correct matches and red for incorrect matches. 
    }
    \label{fig:offline_qualitative}
\end{figure}

\begin{figure}
    \centering
    \includegraphics[width=\textwidth]{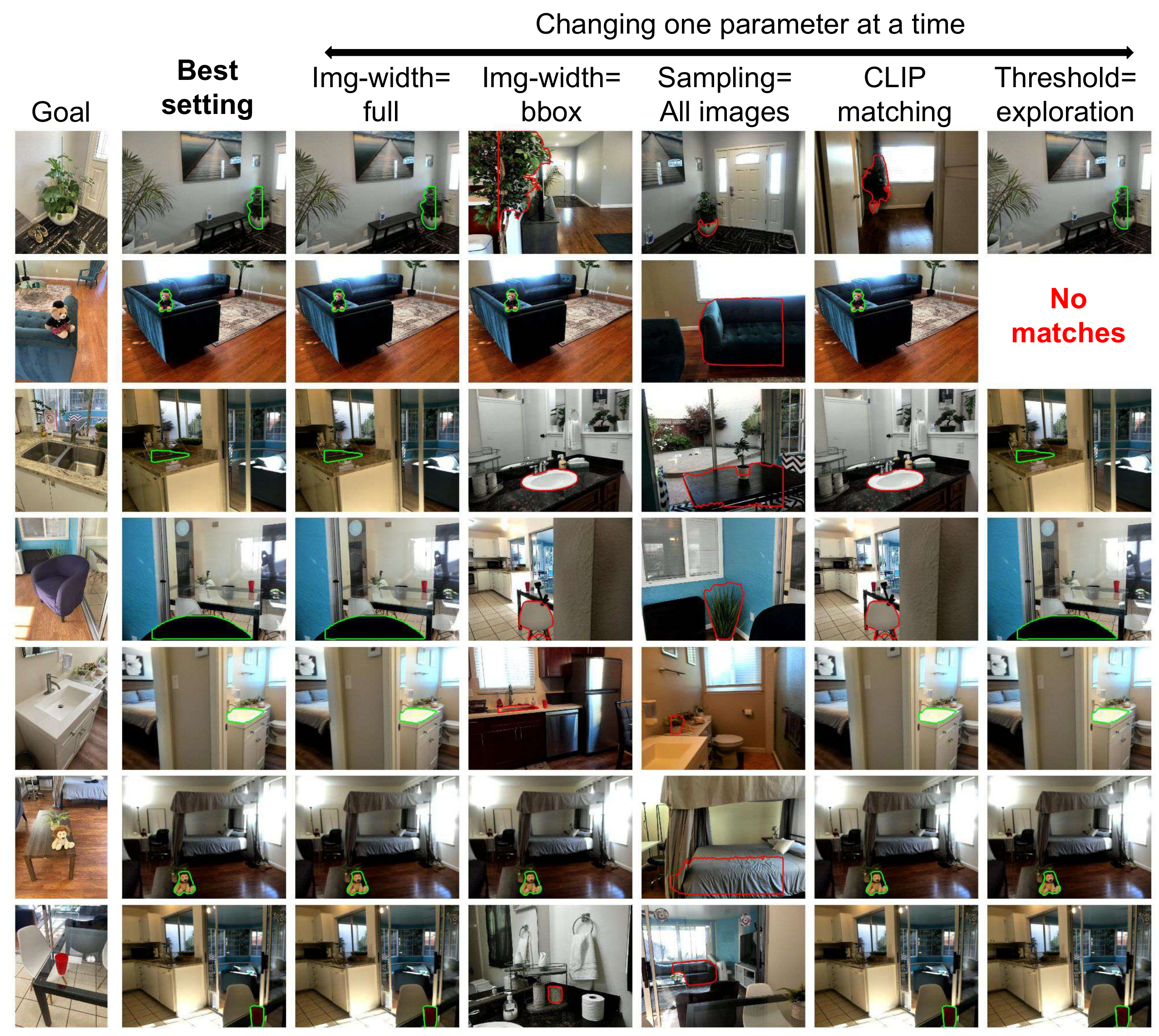}
    \caption{
    \textbf{Matching strategy ablation for selecting image goals.}
    In the second column, we show the image goals identified by our best setting (Img-width=Bbox+pad, Agg-fn=max, Sampling=By Category, Threshold=0.0, Method=SuperGLUE). In the subsequent columns, we show results after changing one parameter at a time with respect to the best setting. The object instance selected in each setting is highlighted with a border which is green for correct matches and red for incorrect matches.
    }
    \label{fig:enter-label}
\end{figure}

\clearpage
\bibliographystyle{plain}
\bibliography{scibib}

\end{document}